\documentclass[journal]{IEEEtai}

\usepackage[colorlinks,urlcolor=blue,linkcolor=blue,citecolor=blue]{hyperref}

\usepackage{color,array}
\usepackage{multirow}
\usepackage{graphicx}
 \usepackage{amsmath}
 \usepackage{amssymb}

\begin{document}

\title{Convolutional vs Large Language Models for Software Log Classification in Edge-Deployable Cellular  Network Testing}

\author{Achintha Ihalage*, Sayed M. Taheri*, Faris Muhammad, Hamed Al-Raweshidy


\thanks{Achintha Ihalage and Faris Muhammad are with the Wireless Business Unit of VIAVI Solutions Inc., United Kingdom. (e-mail: \{achintha.ihalage and faris.muhammad\}@viavisolutions.com).}

\thanks{Sayed M. Taheri, is with the Department of Electronic and Electrical Engineering (EEE), College of Engineering, Design and Physical Sciences, Brunel University of London, London, United Kingdom. Since 2022, he has also been with VIAVI Solutions Inc., United Kingdom. (e-mail: Sayed.Taheri@Brunel.ac.uk, and, Sayed.Taheri@viavisolutions.com).}

\thanks{Hamed Al-Raweshidy \textit{(Corresponding author)} is with the Wireless Networks and Communications Group, Department of   Electronics and Electrical Engineering, Brunel University of London, London, United Kingdom. (e-mail: Hamed.Al-Raweshidy@Brunel.ac.uk).}

\thanks{*These authors contributed equally.}
}


\maketitle

\begin{abstract}

Software logs generated by sophisticated network emulators in the telecommunications industry, such as VIAVI TM500, are extremely complex, often comprising tens of thousands of text lines with minimal resemblance to natural language. Only specialised expert engineers can decipher such logs and troubleshoot defects in test runs. While AI offers a promising solution for automating defect triage, potentially leading to massive revenue savings for companies, state-of-the-art large language models (LLMs) suffer from significant drawbacks in this specialised domain. These include a constrained context window, limited applicability to text beyond natural language, and high inference costs. To address these limitations, we propose a compact convolutional neural network (CNN) architecture that offers a context window spanning up to 200,000 characters and achieves over 96\% accuracy (F1$>$0.9) in classifying multifaceted software logs into various layers in the telecommunications protocol stack. Specifically, the proposed model is capable of identifying defects in test runs and triaging them to the relevant department, formerly a manual engineering process that required expert knowledge. We evaluate several LLMs; LLaMA2-7B, Mixtral\_8x7B, Flan-T5, BERT and BigBird, and experimentally demonstrate their shortcomings in our specialized application. Despite being lightweight, our CNN significantly outperforms LLM-based approaches in telecommunications log classification while minimizing the cost of production. Our defect triaging AI model is deployable on edge devices without dedicated hardware and widely applicable across software logs in various industries.

\end{abstract}

\begin{IEEEImpStatement}
Complex raw logs generated by software and hardware stacks prevalent in the telecommunications industry pose unique challenges in understanding and diagnosing relevant issues. Due to the nature of such logs, it becomes crucial to analyze their entire content to identify defects based on network configurations, dynamic parameters, and combinations of log messages. Existing machine learning-based methods that support only narrow input sequences of text prove inadequate in this situation. This article introduces a robust convolutional neural network (CNN) architecture that supports input text sequences up to 200,000 characters, enabling effective learning from voluminous software logs. An experimental study is conducted to illustrate the effectiveness of tailored machine learning architectures in specialized domains by benchmarking the proposed model against recent large language models. In addition to delivering superior performance in software log classification, our model is practically feasible to be deployed in industrial settings.


\end{IEEEImpStatement}


\section{Introduction}


As the telecommunications industry rapidly evolves into the 5G and 6G era, the scope and complexity of network testing have expanded at an unprecedented rate. Such testing solutions entail emulating user equipment (UE) and their intricate interactions with the network infrastructure. This enables network operators and equipment manufacturers to evaluate the quality and capacity of their wireless networks, ensure compliance with industry-standard protocols, and effectively troubleshoot issues while optimizing network performance. Meeting these needs involves employing advanced dedicated hardware (such as VIAVI TM500). 

Invariably, these sophisticated systems generate voluminous and highly complex software logs that are indispensable for troubleshooting and defect triaging. The telecom raw logs analyzed in this context stand apart in their complexity and nature from conventional software logs, exhibiting a vast diversity of commands and parameters configured in real-time under dynamic industrial conditions. Moreover, their structure and semantics bear little correlation to natural language.  Therefore, only expert engineers with extensive experience in the field can navigate through these logs to identify issues in a network emulation. However, this manual analysis is inefficient, error-prone, less scalable, and vulnerable to knowledge silos. This inefficiency in manual analysis is further compounded by the high stakes involved; delays in defect resolution can have significant financial and reputational repercussions for both service providers and their clients.

 Recently, the application of machine learning (ML) and natural language processing (NLP) techniques has demonstrated remarkable success in the classification and extraction of insights from text-based software logs \cite{ulku_2021, Catovic_2021, Pitakrat_2015, Aljabri_2022, Voloskin_2022, Abdalla_2022, Ren_2018, Meng_2020}. Additionally, a multitude of research studies have been dedicated to log analysis using ML techniques for anomaly detection,  proposing ML solutions ranging from supervised classical models \cite{Bertero_2017, Meng_2021_partial_lbls} to convolutional neural networks (CNNs) \cite{Lu_2018}, recurrent neural network (RNNs) and long short-term memory (LSTM) based networks or hybrid models \cite{Vinayakumar_2017, Yadav_2020}, large language models (LLMs) \cite{Lee_2023, Ott_2021}, other transformer-based models \cite{Huang_Hitanomaly_2020} and unsupervised approaches \cite{Ergen_2019}. Classical ML approaches commonly regard text input as static, relying on a global input text embedding, whereas CNNs, LSTMs, and LLMs handle text by processing it as a sequence of word or token embeddings.

Nevertheless, studies that employ ML for industrial telecommunications logs classification are few and far between. Closest related work \cite{ulku_2021} explores term frequency-inverse document frequency (TF-IDF) and bag of words (BOW) methods for representing software log lines in voice over internet protocol (VoIP) soft-switch systems and classifying them using various classical ML algorithms such as random forest (RF), support vector machine (SVM) and boosting methods. The objective of this study is to distinguish specific log entries into the error, debug, or info classes rather than diagnosing underlying root causes indicated by the logs. Furthermore, Ramachandran \textit{et al}. \cite{Ramachandran_2023} propose a CNN-LSTM architecture to categorize individual lines in industrial telecom-related log files as either errors or non-errors. However, it does not consider the entirety of the software logs for classification. In scenarios like ours, where it is essential to aggregate all errors, warnings, and indications within the logs to identify issues, such methods may not capture the full context necessary for accurate problem determination. In fact, none of the prior studies in the realm of log classification has demonstrated the capability to classify massive software logs containing tens of thousands of text lines while taking into account their complete content. 

It is worthwhile noting the drawbacks of state-of-the-art (SOTA) LLMs in our specific application, despite their remarkable success in text classification \cite{Devlin_BERT_2018}, text generation \cite{Brown_GPT3_2020}  and many other NLP tasks \cite{Zhao_2023, Bariah_2023, Singhal_2023}. LLMs are primarily pretrained and fine-tuned for natural language understanding, including code. Their adaptability to novel text structures, like our proprietary logs, remains constrained due to limited prior exposure to such formats. Therefore, it is essential to pretrain off-the-shelf models on our own extensive software log corpus before fine-tuning them for a log classification downstream task to achieve optimal performance. However, this requires enormous amount of data (e.g., 20 times the number of parameters in the model to achieve compute-optimal state  according to recent research \cite{Hoffmann_chinchilla_2022}) and computing power, hampering the adaptability of LLMs to our use case. Moreover, LLMs typically operate with a relatively narrow context window, ranging from 512 to 2048 tokens. This window is considerably smaller than the volume of text contained in our software logs, resulting in truncation and potential loss of critical information. Recent findings indicate that even the long-context LLMs may exhibit suboptimal performance, particularly when relevant information is situated in the middle of the context window rather than at the beginning or end \cite{Liu_lost_middle_2023}. Lastly, LLMs demand significant memory and computational resources even for inference, which may be undesirable in industrial settings due to higher recurring costs in model serving and deployment.

In this work, we propose a compact residual CNN architecture designed for the classification of software logs produced by VIAVI TM500. Its primary function is to identify the specific protocol stack layer in which a defect has occurred, thereby facilitating a more streamlined defect triage process. Our model can process text sequences up to 200,000 characters at a time and achieves over 96\% accuracy (F1\_score $>$ 0.9) in classifying complex software logs. To construct an initial embedding matrix, we trained a separate LSTM-based model with a sequence-to-sequence (seq2seq) objective on our sufficiently-sized text corpus. This process is analogous to \textit{software log language} understanding and inspired by recent work that underscores the importance of industry-specific word embeddings in downstream log classification tasks \cite{Khabiri_2019}. Finally, we benchmark the performance of our CNN against latest open-source LLMs, namely Mixtral\_8x7B, LLaMA2\_7B \cite{Touvron_llama2_2023}, Flan-T5 \cite{Chung_flant5_2022}, BigBird \cite{Zaheer_bigbird_2020} and BERT \cite{Devlin_BERT_2018}.  

The main contributions and novelty of this article are as follows.

\begin{enumerate}
    \item We introduce a lightweight residual CNN architecture that can classify massive software logs accurately for defect detection in 5G/6G cellular network testing. To the best of our knowledge, the proposed ML model is the first of its kind that can support input text sequences up to 200,000 characters in telecommunications domain, with broad implications for software log classification across disciplines. Our model is only 3 Mega-bytes in size and has less than 0.8 Million parameters, making it edge-deployable and production-ready.

    \item We experimentally demonstrate the drawbacks of off-the-shelf LLMs in specialized applications. To tailor these models for our specific domain, we perform domain-specific pretraining and subsequent finetuning on our data using techniques such as Low-Rank Adaptation (LoRA) \cite{Hu_lora_2021} and quantization where applicable.

    \item We propose an adaptable overlapping sliding window approach to extract pretrained-LLM embeddings of lengthy software logs that often exceed the context window of LLMs. Embeddings are extracted from several LLMs, including the latest models such as Mixtral\_8x7B LLaMA2\_7B that contain Billions of parameters. Separate classifiers are applied on top of these embeddings, showcasing acceptable classification performance.

    \item We develop a sequence-to-sequence model based on LSTM units for understanding raw logs in telecommunications. Meaningful token embeddings extracted from this model are utilized to initialize the Embedding layer of the proposed CNN, helping to achieve optimal performance. 

\end{enumerate}

The remainder of this article is organized as follows. Section \ref{sec:related_work} presents an overview of the literature in the field of log classification and log anomaly detection. Section \ref{sec:methods} elucidates the proposed ML architectures with theoretical foundations and provides a comprehensive overview of our data processing pipeline. In Section \ref{sec:results_and_discussion}, we present the results and analysis of our model and conduct a benchmarking study with LLMs. Experimental details of model training and evaluation are described in Section \ref{sec:experimental_details}. Finally, Section \ref{sec:conclusion} concludes this article.


\section{Related Work}
\label{sec:related_work}

Analysing software logs has been studied under both supervised and unsupervised learning categories in the literature. Below, we discuss some notable work on log classification and log anomaly detection highlighting key methodologies and their outcomes, which partly inspired our choice of CNN-based architecture for our industrially-motivated telecommunications logs classification use case. 

\subsection{Supervised Methods}

\textbf{Classical ML}

Chen et al. \cite{Chen_2004} presented a decision tree (DT) learning approach to diagnosing failures in large internet sites. They recorded the runtime properties of each request and trained a DT algorithm on the request trace to identify the causes of failures. Their method successfully identified 13 out of 14 true causes of failure in the logs produced by a large internet services system in a production environment, proving its effectiveness in real-world applications. Liang et al. \cite{Liang_2007} developed a methodology for predicting failures in IBM BlueGene/L supercomputers using event logs. They converted event logs into datasets suitable for classification techniques and applied several classifiers, including a rule-based classifier, support vector machines (SVMs), and a customized nearest neighbor method. Their customized nearest-neighbor approach outperformed others in coverage and precision, suggesting its viability in alleviating the impact of failures. One key benefit of some of the classical ML algorithms is their interpretability. For example, a DT may be indicative of which events or log messages are more likely to be related to a failure. Nevertheless, more advanced algorithms have since been proposed, particularly with the progress of deep learning. 

\textbf{RNN and CNN}

Recurrent neural networks, particularly those based on LSTM units, have been extensively applied to log classification due to their capability to handle sequential data and capture long-term dependencies within log sequences. In one of the pioneering studies, Du et al. \cite{Du_deeplog_2017} proposed DeepLog, a deep neural network utilizing LSTM units that can detect anomalies when the log patterns deviate from the normal execution. DeepLog's architecture includes three main components: a log key anomaly detection model, a parameter value anomaly detection model, and a workflow model for diagnosing anomalies. The models are trained with log entries from the normal system execution path. Each log entry is parsed to a log key and parameter value vector, and the models are tasked to predict the probability distribution of the next log key or parameter value vector in their respective sequences. Anomalies are detected at the inference stage by comparing the actual next log key or parameter value with the top predicted ones which model the normal behaviour. If the predictions significantly deviate from the ground truth, as found systematically using thresholds, then the entry is flagged as an anomaly. Zhang et al. \cite{zhang_log_robust_2019} proposed LogRobust, an attention-based bidirectional-LSTM-based neural network for detecting log anomalies in both synthetic public datasets and a real industrial dataset. They model each log event as a semantic vector and use the attention mechanism to weigh different events, as well as a Bi-LSTM architecture to produce anomaly likelihood. The proposed architecture is benchmarked against classical ML models, including SVM and logistic regression, and is demonstrated to achieve consistently high performance. Several other deep learning architectures with LSTM as the key component have been proposed for log anomaly detection \cite{Vinayakumar_2017, Fu_lstm_2023, Xueqing_2021, Yadav_2020}. 

On the other hand, CNN architectures can also be tailored for log classification with a much lower computational load compared to RNN architectures, especially for large context windows. CNNs may work well, especially when the existence of specific log events is important in detecting critical incidents rather than the long-range dependency between the events. A one-dimensional convolutional architecture has been proposed for detecting anomalies in big data system logs such as Hadoop Distributed File System (HDFS) logs \cite{Lu_2018}. Here, the proposed CNN achieved slightly higher performance in relation to a general multi-layer perceptron (MLP) model and an LSTM-based architecture. Ren et al. \cite{Ren_2018} propose a more feature engineering-heavy strategy followed by a two-dimensional CNN to classify CMRI-Hadoop and bluegene/L logs with critical events into 13 different categories. The proposed CNN showcases the best performance among other models, including against classical ML and LSTM-type models.

\subsection{Unsupervised Methods}

Unsupervised log anomaly detection methods can be quite useful, especially when extracting the labels for logs, which is too expensive or impractical. Unsupervised log analysis has been studied using a wide variety of algorithms, from classical models to state-of-the-art transformer-based approaches. Lin et al. \cite{Lin_logcluster_2016} presented LogCluster, an agglomerative hierarchical clustering approach to detect anomalies in Hadoop-based application logs and large-scale online service platforms. Principal component analysis (PCA) has also been applied for log anomaly detection in the absence of a labelled dataset. PCA projects high-dimensional data ($\in \mathbb{R}^N$) into a lower dimensional coordinate system composed of $k$ principal components ($k<N$) that capture the most variance in the original high-dimensional data. Xu et al. \cite{Xu_pca_2009} represented log sequences as event count vectors and used PCA to obtain a lower dimensional representation of logs. They identified anomalies by thresholding the norm of the low-dimensional vectors.

With the remarkable success of the self-attention mechanism and the transformer architecture in language understanding, an increasing number of LLMs have been adapted for log analysis. Lee et al. \cite{Lee_2023} proposed an unsupervised log anomaly detection method based on the BERT architecture \cite{Devlin_BERT_2018}. They first pretrain BERT on normal log data with a masked language modelling (MLM) objective. They then adopt the assumption that the context of a normal system log is notably different from that of an abnormal system log. Under this assumption, a normal log should exhibit a low error and high probability of prediction for masked tokens, whereas an abnormal log may produce a high error and a flatter probability distribution, which allows the detection of anomalies. Almodovar et al. \cite{Almodovar_logfit_2024} presented LogFiT, a finetuned version of the pretrained BERT model for log anomaly detection. They used the masked sentence prediction objective on the normal log data so the model learns the distribution of normal logs. Subsequently, when the model is presented with new log data, its top-k token prediction accuracy can be used with a threshold to identify the anomalous logs. Moreover, several other transformer-based models have been reported for log anomaly detection, including LogBERT \cite{Guo_logbert_2021} and  CCT \cite{Larisch_2023}.

While the literature is rich on general system logs classification and anomaly detection, the space of industrial telecommunications logs classification is largely unexplored. In the following sections, we discuss our approach in classifying logs generated by our network emulation device hardware and associated software stack into various layers in the telecommunications protocol stack.

\section{Methods}
\label{sec:methods}
In this section, we discuss and formulate the proposed models, in particular our seq2seq model for language understanding and the residual CNN for classification.

\subsection{Sequence-to-Sequence Embedding Method}

\begin{figure*}
    \centering
    \includegraphics[width=1\linewidth]{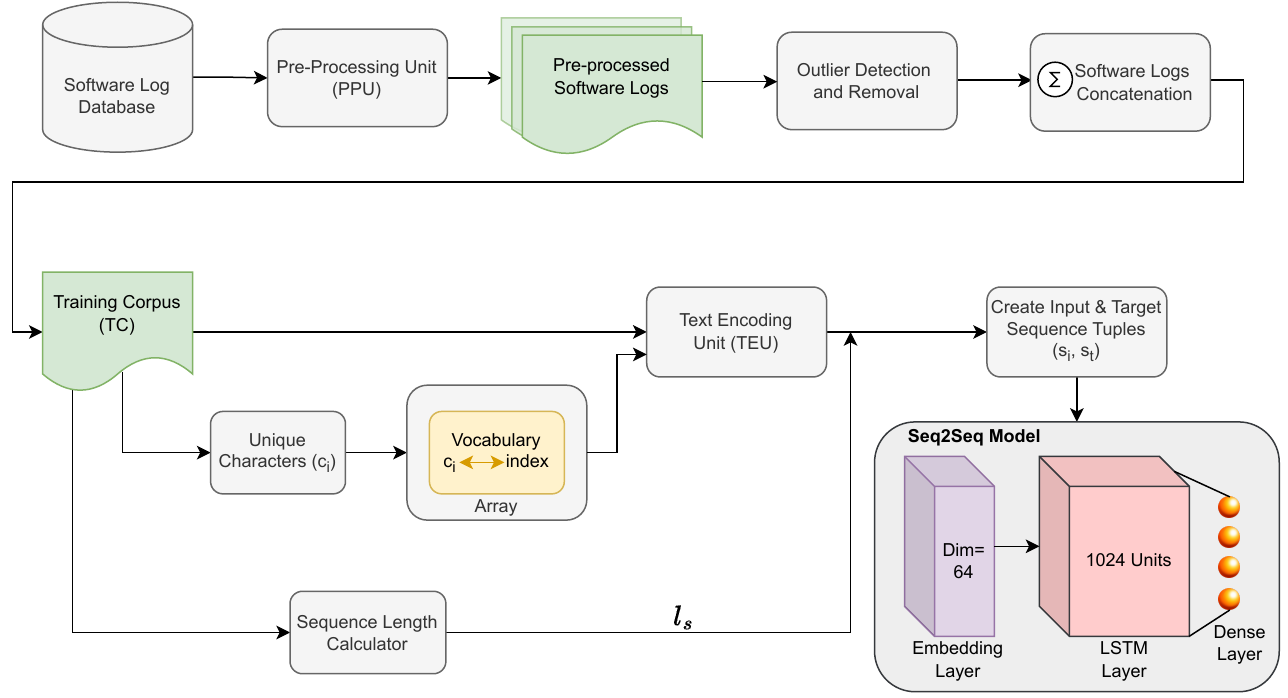}
    \caption{Workflow demonstrating data preparation and architecture of CCOS model.}
    \label{fig:cbos_architecture}
\end{figure*}

\begin{figure}
    \centering
    \includegraphics[width=1\linewidth]{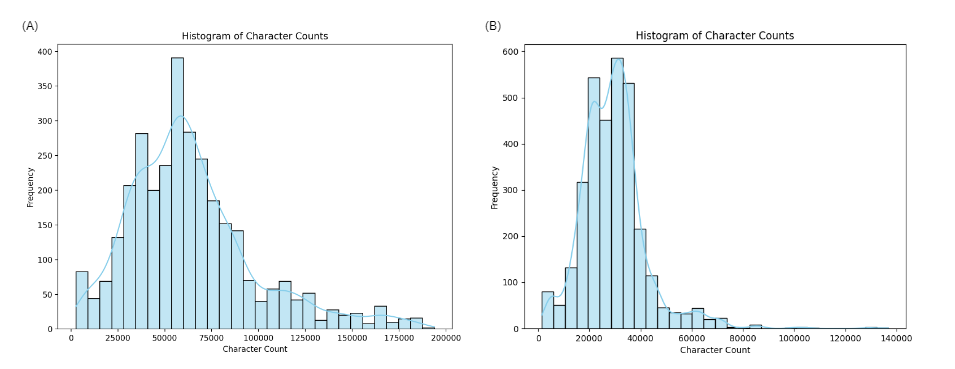}
    \caption{Character length histograms of log files before (A) and after (B) cleaning.}
    \label{fig:char_histogram}
\end{figure}

The objective here is to train a ML model with a seq2seq objective for acquiring learned token embeddings from an industrial corpus. The process of transitioning from a textual representation to a numerical one typically encompasses multiple stages. Fig. \ref{fig:cbos_architecture} shows the steps we follow for preparing the input raw logs into primarily encoded text so that it can be used by our seq2seq model for training. Raw logs are voluminous, often riddled with noise, inconsistencies, and occasional ambiguity. Once a historical raw logs database is collected, we send it through a pre-processing unit (PPU). PPU captures logs related to several network testing categories, namely, single-UE, single-cell, multi-UE, multi-cell, New Radio (NR) 5G tests, Long-Term Evolution (LTE) 4G tests, and L3 tests. Additionally, the PPU removes redundant information unrelated to detecting defects, such as very long words and lines, and numbers. Next, we identify outlier logs based on their size using the Tukey's method. A box plot (see Appendix Fig. \ref{fig:box_plot}) is created, showing the distribution of the number of characters in logs. Outliers are then defined as observations that fall below $Q1 - 1.5*IQR$ or above $Q3 + 1.5*IQR$, where Q1 and Q3 are first and third quartiles, respectively, and IQR is the interquartile range ($Q3-Q1$). These outliers, along with files $>$ 300 kilo-bytes, have been removed to establish our final dataset. Such filtering is necessary as our raw logs can occasionally go up to hundreds of Megabytes in size. Fig. \ref{fig:char_histogram} depicts the histograms of our dataset before and after preprocessing.


\begin{figure*}
    \centering
    \includegraphics[width=1\linewidth]{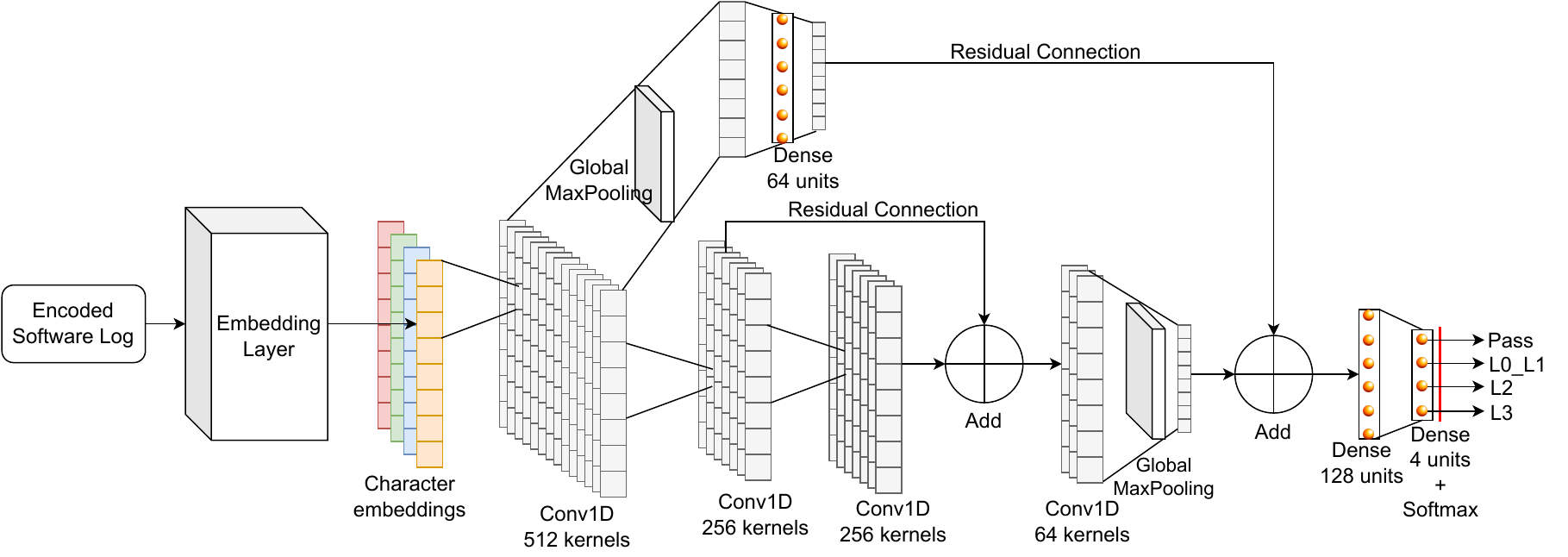}
    \caption{Proposed residual CNN architecture for software log classification.}
    \label{fig:cnn_architecture}
\end{figure*}

The resulting software logs are then concatenated to create a training corpus (TC). Given the little correlation between our logs and natural language, we opted to use unique characters present in our corpus as tokens. This approach yielded in a vocabulary consisting of 97 unique characters, allowing us to encode any piece of text within the TC and obtain a corresponding numerical representation. In order to create training sequences, we choose a sequence length ($l_s$) equal to the median length of \textit{message blocks} (in characters) in our software command logs. A \textit{block} refers to the set of information contained within so-called Indications (\texttt{I:}) that record a set of events that have come back from the network, and Confirmations (\texttt{C:}) that record the state of execution of system commands. These are the information that is critical for the AI system to be able to learn and capture defects, justifying the rationale behind the selected $l_s$. Next, we create tuples of input and target sequences ($s_i$, $s_t$) with matching lengths ($l_s$), where $s_t$ is formed by shifting a window of $l_w$ characters across $s_i$ within the continuous TC. To maintain simplicity, we adopt the assumption that $l_w = 1$.

Having equal and fixed-length input and target sequences allows us to design a simpler recurrent neural network architecture (RNN) without an explicit decoder for seq2seq learning. The proposed architecture as illustrated in Fig. \ref{fig:cbos_architecture} has an embedding layer to represent every character in the vocabulary with an embedding of 64 dimensions, chosen heuristically. This is followed by an LSTM layer with 1024 units that returns processed sequences, and an output fully-connected dense layer with the number of units equal to the vocabulary size. The dense layer is applied across all returned sequences by the LSTM layer. As such, our optimization objective is to minimize the negative log-likelihood of the true next sequence.


\begin{equation}
\label{eq:neg_log}
\text{argmin}_\theta \frac{1}{N} \sum_{j=1}^{N} -\log P(s_j | s_{j-1}; \theta)
\end{equation}

where N is the total number of input and target sequence tuples in the dataset. $P(s_j | s_{j-1}; \theta)$ is the conditional probability of generating the true next sequence $s_j$ given the previous sequence $s_{j-1}$ and the model parameters $\theta$. With this notation, the target sequence $s_t$ equals to $s_j$, and the input sequence $s_i$ equals to $s_{j-1}$. Equation \ref{eq:neg_log} can be further decomposed as follows, considering individual characters.

\begin{equation}
\label{eq:char_neg_log}
   \text{argmin}_\theta \frac{1}{N} \sum_{j=1}^{N} \sum_{t=1}^{l_s} -\log P(c_t^{s_j} | c_1^{s_{j-1}}, ..., c_{l_s}^{s_{j-1}}; \theta)
\end{equation}

where $c_t^{s_j}$ is the character at the $t^{th}$ index of sequence $s_j$. Probability logits over the vocabulary for every character in the target sequence are calculated through a softmax function applied at the output layer. Consequently, the model is trained to minimize the multi-class categorical cross-entropy loss. Once the model is trained, character embeddings are extracted from the weights stored within the Embedding layer of the model. It is worth noting that the trained seq2seq model, primarily utilized for extracting character embeddings, has the potential for software log text generation. However, this aspect falls outside the scope of this paper.

\subsection{Residual CNN for Software Log Classification}

The proposed architecture for lengthy software log classification is primarily based on one-dimensional convolutional layers (Conv1D). CNNs are generally lightweight and consume less computational resources. This architectural choice stems partly from practical considerations in industrial production, as many target edge devices lack dedicated hardware like GPUs.  Nevertheless, as we demonstrate later in this article, the proposed CNN, in fact, consistently outperforms other benchmarked models. Formally, Conv1D operation applied on a discrete 1D input sequence $s$ at a time index $t$ with single filter $w$ is given by;

\begin{equation}
\label{eq:conv_operation}
y(t) = (w * s)(t) = \sum_{k=-\frac{(K-1)}{2}}^{\frac{(K-1)}{2}} w(k) \cdot s(t-k)
\end{equation}

where $*$ indicates the convolution operation, $y(t)$ is the feature map resulting from the filter applied at position $t$, and $K$ is the kernel size.

Concretely, our CNN consists of an embedding layer initialized with character embeddings extracted from the seq2seq model, followed by a block of Conv1D layers before applying global max pooling operation on the processed sequences, resulting in a fixed-size representation of the input sequence. This is then processed through a set of fully-connected layers followed by softmax activation for multi-class classification. We further apply residual connections to improve model performance. The end-to-end (E2E) architecture is depicted in Fig. \ref{fig:cnn_architecture}. 

\begin{figure}
    \centering
    \includegraphics[width=1\linewidth]{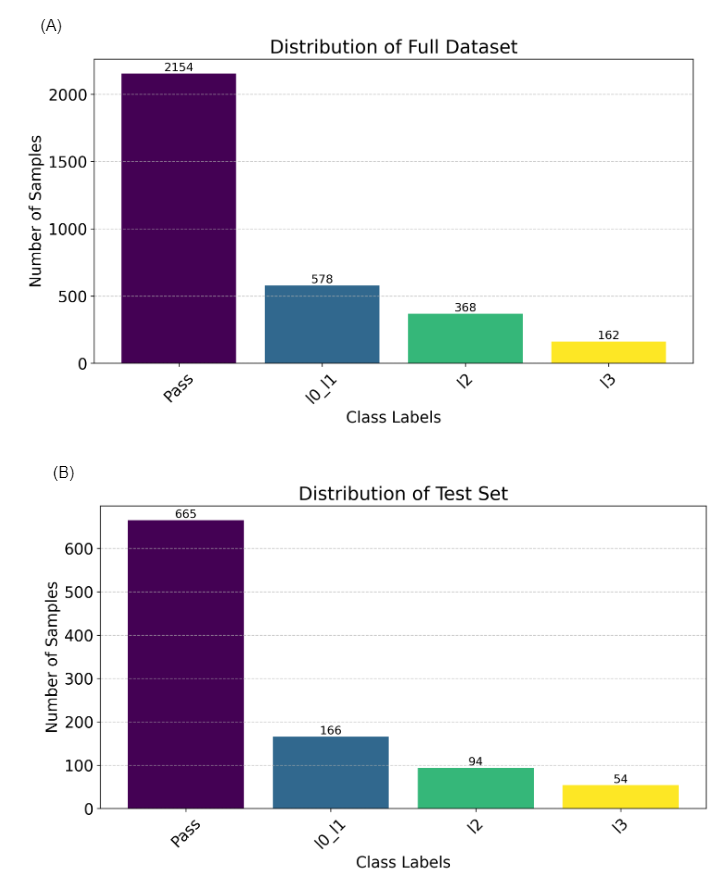}
    \caption{Distribution of classes in our full dataset (A) and test set (B).}
    \label{fig:dataset_distribution}
\end{figure}

The model is trained to classify various software logs into four distinct classes: $Pass$, $L0\_L1$, $L2$, and $L3$. $Pass$ class represents software logs that do not indicate any issues. $L0\_L1$ denotes defects at the physical layer, $L2$ pertains to issues within the data link layer, and $L3$ encompasses problems related to the network or higher layers, in accordance with the Open Systems Interconnection (OSI) model. These labels for the software logs in our dataset are extracted from historical data. Undoubtedly, most test runs are completed without issues, resulting in a highly imbalanced class distribution within our dataset, as can be seen in Fig. \ref{fig:dataset_distribution}(A). To reduce the impact of class imbalance, ML models studied in this work are trained with appropriate class weights where applicable. There are 3262 samples in total in our dataset which is randomly divided into 70\% training and 30\% test sets. The class distribution for the test set is shown in Fig. \ref{fig:dataset_distribution}(B).

\section{Results and Discussion}
\label{sec:results_and_discussion}
In this section, we present the classification results of the proposed CNN on the test set and benchmark it against several other approaches, including LLMs. Note that the test set is kept intact across all experiments. 

\subsection{CNN Results}

\begin{figure*}
    \centering
    \includegraphics[width=1\linewidth]{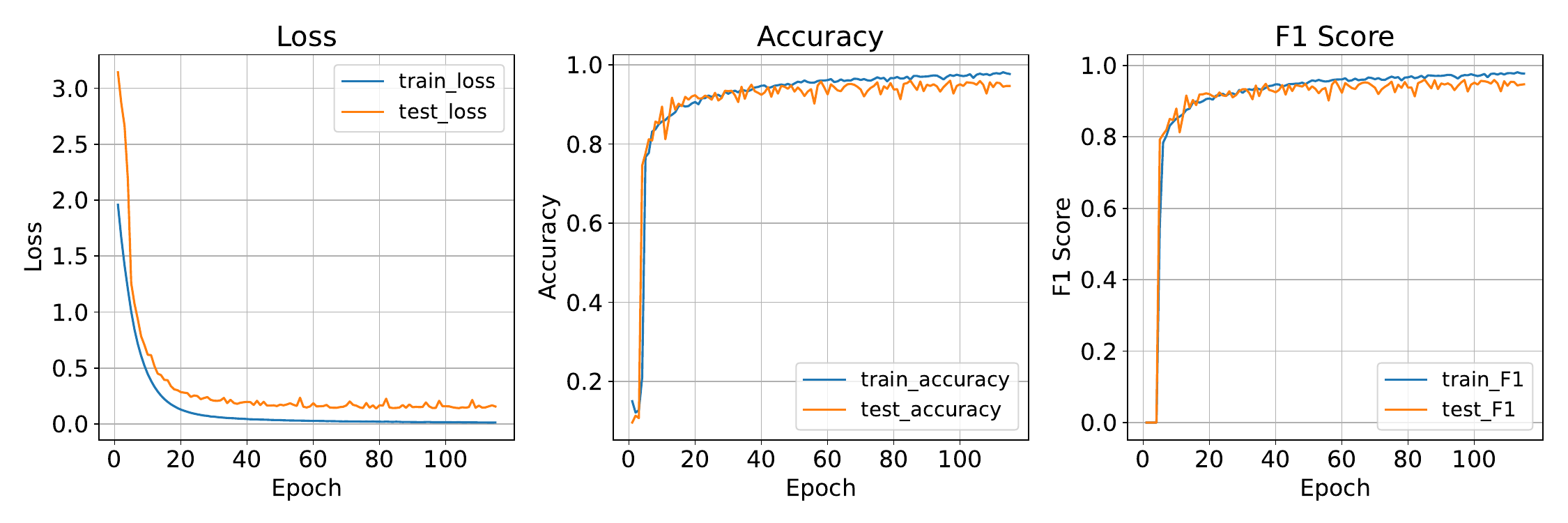}
    \caption{Training progress curves for the proposed residual CNN model. Note that the F1-score shown here is the micro average computed by obtaining the overall true positives, false positives, and false negatives across all classes.}
    \label{fig:training_curves}
\end{figure*}

Fig. \ref{fig:training_curves} illustrates the training progress of our CNN, namely the categorical cross-entropy loss, accuracy, and F1-score curves.  Our defect triaging model achieves a notable 96\% accuracy, markedly improving both precision and efficiency in manual engineering workflows.  A more comprehensive summary of results is shown in Table \ref{tab:cnn_bilstm_results}. Although our model can support text sequences up to 200,000 characters, we used a sequence length of 50,000 characters to obtain results shown in Table \ref{tab:cnn_bilstm_results}. This is because nearly 95\% of our cleaned software logs are less than 50,000 characters in length. The trained model, with its compact size of just 3 megabytes, is ideally suited for deployment on edge devices with basic hardware resources.

\begin{figure*}
    \centering
    \includegraphics[width=1\linewidth]{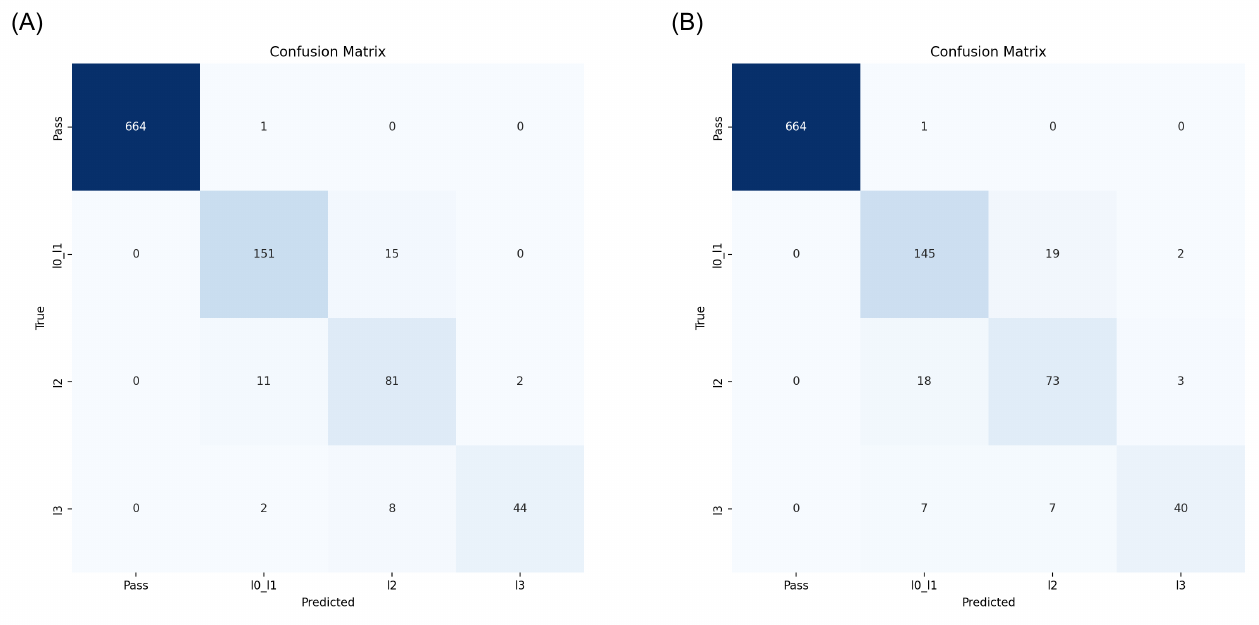}
    \caption{Confusion matrices of CNN (A) and CNN+BiLSTM (B) models for the test set.}
    \label{fig:conf_mat}
\end{figure*}

On the other hand, LSTM networks are recognised for their effectiveness in sequence classification tasks, owing to their ability to learn long-term dependencies with memory cells and gating mechanism. It is interesting to include LSTM components in our model and assess the impact on performance. In fact, such architectures have been proposed for sequence learning tasks in telecommunications including channel estimation \cite{Nguyen_2023}.We added a bidirectional LSTM layer right after the embedding layer of the proposed CNN (denoted as BiLSTM+CNN model) and evaluated the performance on the test set under the same conditions. This resulted in a slightly downgraded accuracy of 94.2\%, despite a nearly three-fold increase in model size (see Table \ref{tab:cnn_bilstm_results}).  This potentially suggests that in the context of our software logs, it is not so much the long-range dependencies or the semantic structure that are crucial for identifying defects but rather the presence of specific combinations of log messages. The confusion matrices of the CNN and CNN+BiLSTM models are displayed in Fig. \ref{fig:conf_mat}. Both models can accurately detect even the least represented class ($L3$) that has less than one-tenth of samples compared to the dominant class.

\begin{figure}
    \centering
    \includegraphics[width=1\linewidth]{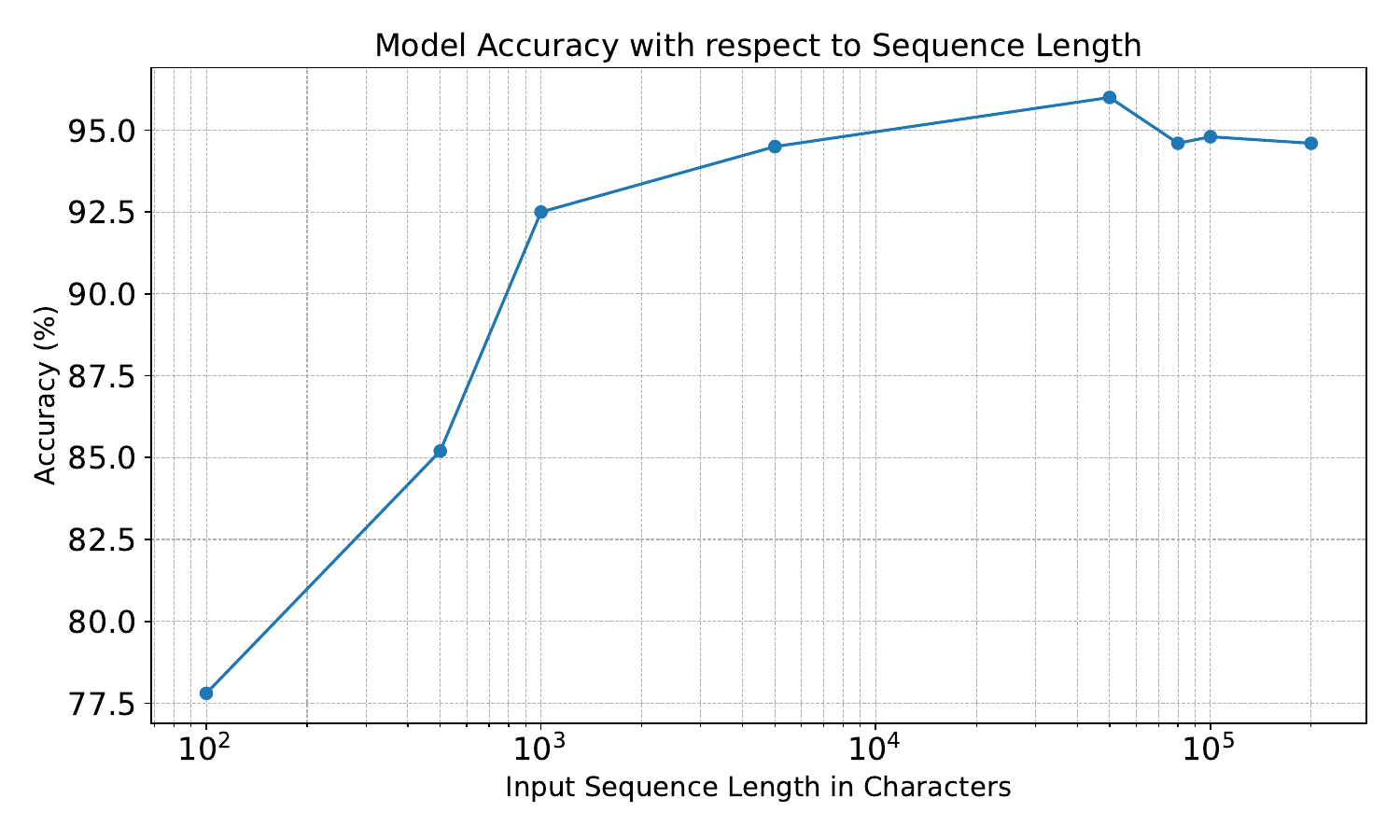}
    \caption{Performance of residual CNN with varying context size. Note that the x-axis is in the log scale.}
    \label{fig:acc_vs_char_len}
\end{figure}

We subsequently explored how the input sequence length impacts the performance of our CNN. Illustrated in Fig. \ref{fig:acc_vs_char_len}, we observed a steady improvement in performance as the sequence length increases, up until a point where it begins to marginally decline for extremely lengthy sequences ($>$80k characters). This trend may occur because such expansive context windows necessitate excessive padding in most input text sequences with redundant tokens, thereby diluting the pertinent information. Furthermore, we noticed that the proposed CNN model is generally robust in terms of the number of Conv1D layers present in the architecture. Employing just a single Conv1D layer while maintaining the rest of the architecture (embedding \& dense layers) intact yielded an accuracy of 93.6\%. Further increasing the number of Conv1D layers, up to four, repeatedly led to a classification accuracy of above 95\% with consistent distribution of other performance metrics.

\begin{table*}[]
\centering
\caption{Multi-class performance metrics of the residual CNN and BiLSTM+CNN models on the test set.}
\label{tab:cnn_bilstm_results}
\begin{tabular}{|c|c|c|c|cccc|}
\hline
\multirow{2}{*}{Model}      & \multirow{2}{*}{\# Parameters} & \multirow{2}{*}{\begin{tabular}[c]{@{}c@{}}Overall\\ Accuracy (\%)\end{tabular}} & \multirow{2}{*}{\begin{tabular}[c]{@{}c@{}}Overall\\ F1-score (macro)\end{tabular}} & \multicolumn{4}{c|}{Per-Class Metrics}                                                                \\ \cline{5-8} 
                            &                                &                                                                                    &                                                                                       & \multicolumn{1}{c|}{Class}  & \multicolumn{1}{c|}{F1-score} & \multicolumn{1}{c|}{Precision} & Recall \\ \hline
\multirow{4}{*}{CNN}        & \multirow{4}{*}{0.79M}         & \multirow{4}{*}{\textbf{96.01}}                                                             & \multirow{4}{*}{\textbf{0.902}}                                                                & \multicolumn{1}{c|}{Pass}   & \multicolumn{1}{c|}{0.999}    & \multicolumn{1}{c|}{1}         & 0.998  \\ \cline{5-8} 
                            &                                &                                                                                    &                                                                                       & \multicolumn{1}{c|}{L0\_L1} & \multicolumn{1}{c|}{0.912}    & \multicolumn{1}{c|}{0.915}     & 0.91   \\ \cline{5-8} 
                            &                                &                                                                                    &                                                                                       & \multicolumn{1}{c|}{L2}     & \multicolumn{1}{c|}{0.818}    & \multicolumn{1}{c|}{0.779}     & 0.862  \\ \cline{5-8} 
                            &                                &                                                                                    &                                                                                       & \multicolumn{1}{c|}{L3}     & \multicolumn{1}{c|}{0.88}     & \multicolumn{1}{c|}{0.957}     & 0.815  \\ \hline
\multirow{4}{*}{BiLSTM+CNN} & \multirow{4}{*}{2.1M}          & \multirow{4}{*}{94.2}                                                              & \multirow{4}{*}{0.856}                                                                & \multicolumn{1}{c|}{Pass}   & \multicolumn{1}{c|}{0.999}    & \multicolumn{1}{c|}{1}         & 0.998  \\ \cline{5-8} 
                            &                                &                                                                                    &                                                                                       & \multicolumn{1}{c|}{L0\_L1} & \multicolumn{1}{c|}{0.861}    & \multicolumn{1}{c|}{0.848}     & 0.873  \\ \cline{5-8} 
                            &                                &                                                                                    &                                                                                       & \multicolumn{1}{c|}{L2}     & \multicolumn{1}{c|}{0.756}    & \multicolumn{1}{c|}{0.737}     & 0.777  \\ \cline{5-8} 
                            &                                &                                                                                    &                                                                                       & \multicolumn{1}{c|}{L3}     & \multicolumn{1}{c|}{0.808}    & \multicolumn{1}{c|}{0.889}     & 0.741  \\ \hline
\end{tabular}
\end{table*}

\subsection{LLMs for Software Log Classification}

While LLMs have been generally successful in learning to categorize software logs \cite{Huang_Hitanomaly_2020, Lee_2023}, raw logs generated by software and hardware stacks prevalent in the telecommunications industry pose unique challenges to LLMs, particularly due to their vast size and little relevance to the natural language. It may be possible to intuitively reason that the small context windows of off-the-shelf LLMs cannot possibly capture all necessary information from large logs, which may lead to poor downstream performance. Similarly, conducting further pretraining (also known as domain adaptation) on a domain-specific corpus could be beneficial in enhancing downstream task accuracy. We investigate five pretrained LLMs: LLaMA2-7B, Mixtral\_8x7B, Flan-T5, BERT, and BigBird in our specific application. 

While every model undergoes evaluation in its original pretrained state, both Flan-T5 and BERT are additionally chosen for domain adaptation, owing to their distinctive natures and more manageable sizes. Next, we discuss the exact steps that we follow for pretraining LLMs on our own industrial software log corpus. 

\textbf{Domain adaptation} or further \textbf{pretraining} of LLMs may significantly elevate the downstream task performance of LLMs, especially in domains where the morphology of text is considerably different from the natural language on which most models are trained on. First, we identify some specific pieces of text, such as hexadecimal strings, standalone numbers, IP addresses, and new-line characters, that are generally unrelated to identifying defects. These are then masked with special tokens such as \texttt{<hex>}, \texttt{<num>}, \texttt{<ipaddr>} and \texttt{<newline>}. Furthermore, the end of a software log sample in our concatenated training corpus is indicated with an \texttt{<endsample>} token. Tokenization is a critical step in preparing data for LLM training that involves breaking down raw text into numerical representations. The models investigated in this study use subword tokenization, which balances the need to represent common words as they appear and decompose less common ones into smaller, understandable sub-units, resulting in efficient handling of out-of-vocabulary words. Here, we use the pre-trained WordPiece tokenizer for BERT and the SentencePiece tokenizer for Flan-T5. The special tokens identified above are added to both tokenizers before tokenizing the TC. 

We then train BERT for domain adaptation with a warm start from its pre-trained base model checkpoint. The relatively small size of BERT allows for updating all model weights during domain adaptation, conducted with a masked language modeling (MLM) objective on the same software log TC constructed in the workflow shown in Fig. \ref{fig:cbos_architecture}, further processed to add special tokens discussed above. Note that one of the pre-training strategies of BERT, next sentence prediction (NSP), is not used here as performing domain adaptation using MLM is deemed sufficient \cite{Zaheer_2020}. In MLM, a portion of tokens are randomly masked ($\sim$30\% in our case), and the model is tasked to predict these masked tokens based on their context, enabling it to gain a deep understanding of the syntax, semantics, and morphology of a given text corpus.

Flan-T5 is trained starting from \texttt{flan-t5-large} checkpoint using a sequence-to-sequence objective where the target sequences are created by shifting the input sequences
by $k$ ($\geq1$) number of tokens. TC is then tokenized with Flan-T5's own pre-trained tokenizer, extended to include our special tokens. Due to the large size of Flan-T5 architecture ($\sim$780M parameters), we opted to use LoRA method for domain adaptation.  LoRA involves introducing trainable low-rank matrices to modify existing weights in selected layers of the model, such as attention and feed-forward layers, without altering the original pretrained weights. This allows for effective adaptation of the model for specific tasks by reducing the computational burden while maintaining model performance. Both BERT and Flan-T5 are trained for 5 epochs, and the trained checkpoints along with their associated tokenizers are recorded for extracting software log embeddings.

\begin{figure}
    \centering
    \includegraphics[width=1\linewidth]{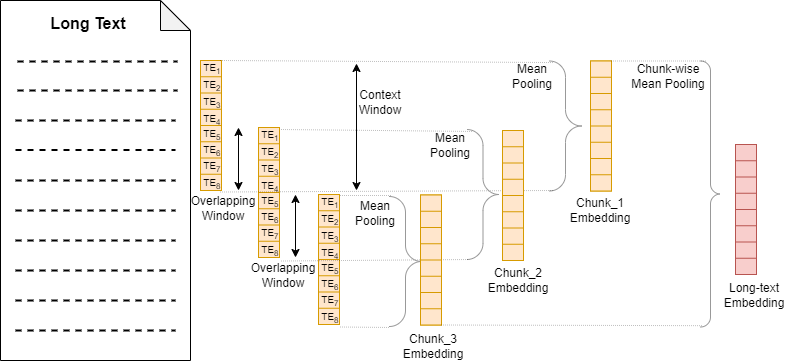}
    \caption{Overlapping sliding window approach to extract embeddings of long documents using LLMs. TE$_i$ refers to the token embedding of $i^{th}$ token at the final hidden layer of the LLM. Token embedding dimension typically ranges from 512 to 4096 for larger models.}
    \label{fig:embedding_extraction}
\end{figure}

\textbf{LLM embedding} of a given text input can be treated as a learned numerical representation that encodes the contextual meaning of that text. As discussed, the small context window of LLMs necessitates an alternative strategy for them to sift through the entirety of long documents. We propose an overlapping sliding window approach, as depicted in Fig. \ref{fig:embedding_extraction}, to extract a single embedding from lengthy text segments. Formally, consider a long text document of length $L$ tokens and a context window (i.e., sequence length) of $l_c$ ($<L$) tokens. For an adaptable overlapping window of size $w$ ($<l_c$) tokens, the number of overlapping text chunks that need to pass through the model to cover the entire document, $M$, can be expressed as follows.

\begin{equation}
\label{eq:n_chunks}
    M = \frac{L-w}{l_c-w}
\end{equation}

Under the proposed approach, the document global embedding $E_{g}$ can be computed as follows.

\begin{equation}
\label{eq:long_emb_calc}
    E_{g} = \frac{1}{M} \frac{1}{l_c} \sum_{k=1}^M  \sum_{i=1}^{l_c} TE_{k,i}
\end{equation}

Here, $TE_{k,i}$ is the token embedding of the $i^{th}$ token of the $k^{th}$ chunk. Token embeddings are extracted at the last hidden layer of the model. $TE_{k,i}$ is of dimension $d_{TE}$ equal to the number of units in the last hidden state of the model. Essentially, the mean-pooling operation is applied across token embeddings of a chunk to obtain the chunk embedding, and the same operation is repeated across all chunks to obtain the document embedding. The attention mask is considered when applying the pooling operation (i.e., pad tokens are disregarded). Text chunks are selected with an overlap $w$ so that some information and context from the previous windows are retained as the model slides through the text. Likewise, embeddings of all logs in our dataset are extracted independently using every LLM examined. Note that due to memory constraints, the context window used for embedding extraction in this work may be smaller than the maximum context window supported by some LLMs. In order to assess the quality of these embeddings, we applied separate classifiers, namely, random forest, XGBoost, and decision tree, with the same defect detection objective. This facilitates benchmarking the proposed LLM-embedding method against our residual CNN. In addition to classical ML models, we utilize a similar residual 1D CNN to classify LLM-embeddings and report the performance on the test set.  

Table \ref{tab: embedding_ML_results} summarizes the experimental results. LLaMA2-7B embeddings used with the XGBoost classifier provide the best accuracy of 82.2\% across all experiments. Although this is reasonable, it represents a notable decline in performance when contrasted with the results achieved by our standalone residual CNN, indicating that even some of the most powerful general-purpose LLMs may perform sub-optimally in challenging domain-specific tasks such as large and complex software log classification. Nevertheless, this series of experiments yields valuable insights. As expected, large and more potent LLMs, specifically, LLaMA2-7B and Mixtral\_8x7B, produce higher quality embeddings, resulting in higher downstream performance compared to other models. Moreover, domain adaptation enhances the performance to some extent, as the models may have gained some understanding of the specific text format by being exposed to a targeted corpus. This enhancement, however, is more prominent for BERT compared to Flan-T5 likely due to distinct domain adaptation strategies. Flan-T5 is pretrained on our corpus by applying LoRA exclusively to the query and value matrices in the transformer block, and therefore the initial token embeddings or model weights that have been optimized for natural language understanding do not get updated. These tokens derived from a corpus of natural language, and their embeddings, may not be very relevant in our specific context. On the contrary, BERT is end-to-end trained during domain adaptation, allowing us to learn token embeddings and model parameters tailored to our corpus, leading to a more substantial boost in performance. Moreover, the downstream task performance is majorly reliant on the quality of embeddings, as more advanced architectures such as 1D-CNNs do not lead to higher performance relative to some classical ML models. 



Finally, we finetune the original pretrained BigBird base model on our labelled dataset with the hope that its relatively large context window of 4096 tokens can capture as much information as possible from software command logs. Instead of extracting embeddings, here, the model is end-to-end finetuned with a classification head, and all model parameters are updated during training. This resulted in accuracy and F1-score of 81\% and 0.581 on the test set, respectively, consistent with LLM-embedding classification results presented in Table \ref{tab: embedding_ML_results}. An analysis of similar-size models reveals that, when the model context window is sufficiently large, finetuning LLMs end-to-end on a downstream classification task leads to higher performance compared to employing separate classifiers on pretrained LLM embeddings. Nevertheless, none of the LLM-based approaches achieves comparable performance to the proposed CNN, indicating that domain-tailored ML architectures may be better suited for industrial applications that require processing miscellaneous formats of text. In addition to delivering superior performance, such lightweight ML models minimize the cost of production and are practically feasible for in-house or field deployment. 

\begin{table*}[]
\footnotesize
\centering
\caption{Classification performance of ML models on the test set using software log embeddings extracted by LLMs. RF and DT refer to random forest and decision tree classifiers, respectively. NA- not applicable.}
\label{tab: embedding_ML_results}
\begin{tabular}{|@{\hspace{1pt}}c@{\hspace{1pt}}|c@{\hspace{2pt}}|c|c|cccc|cccc|}
\hline
\multirow{2}{*}{\begin{tabular}[c]{@{}c@{}}LLM Embedding\\ Model\end{tabular}} & \multirow{2}{*}{\begin{tabular}[c]{@{}c@{}}Approximate\\ \# Parameters\end{tabular}} & \multirow{2}{*}{\begin{tabular}[c]{@{}c@{}}Used Context\\ Window\end{tabular}} & \multirow{2}{*}{\begin{tabular}[c]{@{}c@{}}Embedding\\ Dimension\end{tabular}} & \multicolumn{4}{c|}{Overall Accuracy (\%)}                                                                    & \multicolumn{4}{c|}{Overall F1-score (macro)}                                                                  \\ \cline{5-12} 
                                                                               &                                                                                      &                                                                                &                                                                                & \multicolumn{1}{c|}{RF} & \multicolumn{1}{c|}{DT} & \multicolumn{1}{c|}{XGBoost} & CNN  & \multicolumn{1}{c|}{RF} & \multicolumn{1}{c|}{DT} & \multicolumn{1}{c|}{XGBoost} & CNN   \\ \hline
LLaMA2-7B                                                                      & 7B                                                                                   & 2048                                                                           & 4096                                                                           & \multicolumn{1}{c|}{80.8}          & \multicolumn{1}{c|}{71}            & \multicolumn{1}{c|}{\textbf{82.2}}    & 77.1 & \multicolumn{1}{c|}{0.683}         & \multicolumn{1}{c|}{0.59}          & \multicolumn{1}{c|}{\textbf{0.705}}   & 0.632 \\ \hline
Mixtral\_8x7B                                                                  & 47B                                                                                  & 512                                                                            & 4096                                                                           & \multicolumn{1}{c|}{79.2}          & \multicolumn{1}{c|}{67.2}          & \multicolumn{1}{c|}{80.2}    & 77.4 & \multicolumn{1}{c|}{0.626}         & \multicolumn{1}{c|}{0.517}         & \multicolumn{1}{c|}{0.658}   & 0.617 \\ \hline
Flan-T5                                                                        & 770M                                                                                 & 512                                                                            & 1024                                                                           & \multicolumn{1}{c|}{75.7}          & \multicolumn{1}{c|}{63.9}          & \multicolumn{1}{c|}{79.5}    & NA   & \multicolumn{1}{c|}{0.531}         & \multicolumn{1}{c|}{0.466}         & \multicolumn{1}{c|}{0.649}   & NA    \\ \hline
Flan-T5 (Domain Adapted)                                                       & 770M                                                                                 & 512                                                                            & 1024                                                                           & \multicolumn{1}{c|}{77}            & \multicolumn{1}{c|}{63.3}          & \multicolumn{1}{c|}{79.2}    & NA   & \multicolumn{1}{c|}{0.555}         & \multicolumn{1}{c|}{0.4291}        & \multicolumn{1}{c|}{0.634}   & NA    \\ \hline
BERT                                                                           & 110M                                                                                 & 512                                                                            & 768                                                                            & \multicolumn{1}{c|}{73.9}          & \multicolumn{1}{c|}{62.3}          & \multicolumn{1}{c|}{78.2}    & 74.6 & \multicolumn{1}{c|}{0.45}          & \multicolumn{1}{c|}{0.438}         & \multicolumn{1}{c|}{0.596}   & 0.527 \\ \hline
BERT (Domain Adapted)                                                          & 110M                                                                                 & 512                                                                            & 768                                                                            & \multicolumn{1}{c|}{76.9}          & \multicolumn{1}{c|}{69.1}          & \multicolumn{1}{c|}{79.4}    & 74.6 & \multicolumn{1}{c|}{0.568}         & \multicolumn{1}{c|}{0.507}         & \multicolumn{1}{c|}{0.641}   & 0.568 \\ \hline

BigBird (E2E Finetuned)                                                        & 110M                                                                                 & 4096                                                                           & NA                                                                             & \multicolumn{4}{c|}{81}                                                                                     & \multicolumn{4}{c|}{0.581}                                                                                     \\ \hline
\end{tabular}
\end{table*}

\section{Experimental Details}
\label{sec:experimental_details}

All training and evaluation experiments in this study were performed on KubeFlow, which enables the orchestration of machine learning workflows on the Kubernetes cluster. Ubuntu 20.04 was used as the operating system. The hardware infrastructure consisted of an Nvidia DGX server carrying 8xA100 graphics processing units (GPUs), each with 80GB memory. The system is equipped with 1 terabyte of physical memory. MLFlow \cite{Zaharia_mlflow_2018} integrated within the KubeFlow environment was used for experiment tracking and logging artifacts. 

CNN architecture and hyperparameters were optimized empirically. The model was trained for up to 200 epochs with an early stopping patience of 30 epochs. Adam optimizer with a learning rate of $10^{-4}$ was used to optimize model parameters. $L_2$ regularization was applied to selected layers to reduce overfitting and improve generalization of the model. The batch size was adjusted accordingly within a range of 16 to 512 to accommodate large context sizes. For the largest context size tested (200k), the model completed training in under one hour. A similar hyperparameter setting was employed for the seq2seq LSTM model that had 4.6 Million parameters. We used \textit{Keras} package with \textit{TensorFlow} backend to implement and train the models. 

LLMs analyzed in this study were adapted from Hugging Face \cite{Wolf_2019} using the \textit{transformers}  package. BERT and Flan-T5 models were each trained for 5 epochs for domain adaptation. LoRA adapters, with a rank of 16 and a scaling factor of 32 were used to train Flan-T5. BERT was trained in less than one hour on our TC whereas the training of Flan-T5 took about seven hours. The batch size is set to 64 for BERT and 12 for Flan-T5.  Larger models (LLaMA2-7B and Mixtral\_8x7B) were 4-bit quantized using the \textit{bitsandbytes} package before extracting embeddings for computational efficiency. Flash attention 2 \cite{Dao_2023} implementation in the \textit{transformers} package was used to further reduce memory requirements. The overlapping window size was set to half the sequence length for embedding extraction. \textit{torchrun} was used for distributed training of LLMs where applicable. BigBird base model was finetuned for up to 200 epochs with a batch size of 8 and an early stopping patience of 30 epochs. In general, the finetuning of BigBird took about 4 hours.  The hyperparameters of the ML models used with LLM-embeddings were optimized using a 5-fold cross-validation strategy with grid search. Classical ML implementations were adopted from Scikit-Learn \cite{Pedregosa_2011} with default parameters. Classification performance metrics are calculated as follows where TP, FP and FN indicate true positives, false positives and false negatives, respectively.

\begin{equation}
    \text{Precision} = \frac{TP}{TP + FP}
\end{equation}

\begin{equation}
    \text{Recall} = \frac{TP}{TP + FN}
\end{equation}

\begin{equation}
    \text{F1\_score} = 2 \times \frac{\text{Precision} \times \text{Recall}}{\text{Precision} + \text{Recall}}
\end{equation}

\section{Conclusion}
\label{sec:conclusion}
In this research, we presented a robust CNN architecture tailored for the classification of intricate large-scale software logs in the telecom sector to meet the edge-deployability requirement of defect detection systems. The proposed model, adept at processing extensive text sequences up to 200,000 characters, markedly outperforms some LLM-based approaches examined. This advancement is critical in mitigating the limitations of manual log analysis, such as inefficiency and error-proneness, offering a streamlined, automated approach for defect triage in 5G/6G network testing. Our research demonstrates that custom, compact models like our CNN architecture can not only offer a practical and efficient alternative to resource-intensive LLMs in specific industrial applications, but can even surpass their performance. While
our results do not establish that custom CNN architectures
can universally surpass LLMs in all log classification tasks,
this study underscores the importance of considering architectural choices in light of real-world constraints and production overheads.
In conclusion, the present study not only contributes a novel, edge-deployable ML model for accurate defect detection from raw software logs in the telecommunications industry but also provides valuable insights into the application of artificial intelligence in industrial settings, paving the way for future innovations in software log classification.



\section*{Appendix}

\begin{figure}
    \centering
    \includegraphics[width=1\linewidth]{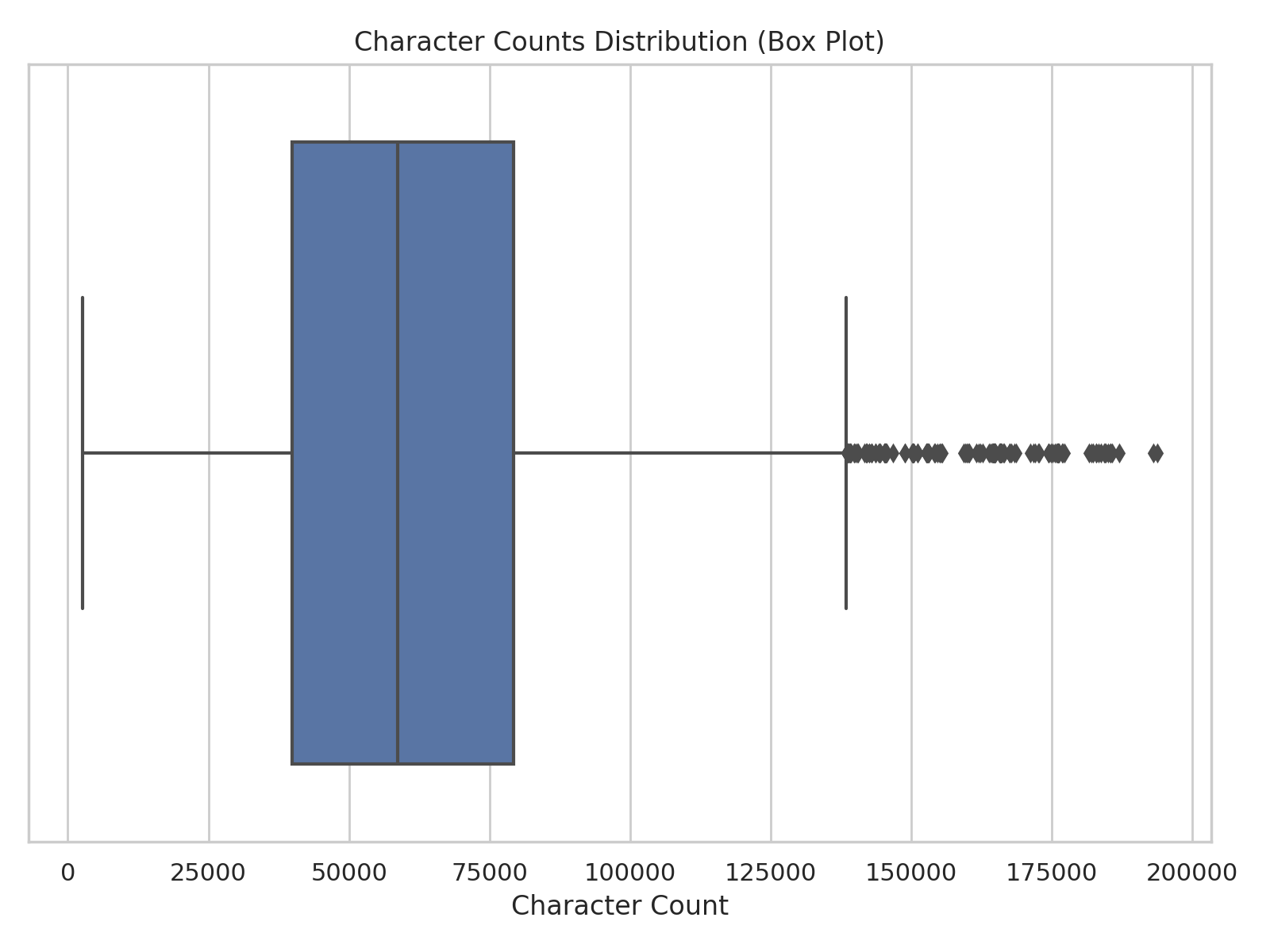}
    \caption{Boxplot indicating outlier software logs.}
    \label{fig:box_plot}
\end{figure}


\section*{Acknowledgment}

We acknowledge the support of VIAVI Solutions Inc. for their provision of data, funding, and MLOps infrastructure, including GPUs, which contributed significantly to the completion of this project.



\section*{References}



\def\refname{}

\end{document}